\newcommand{\CLASSINPUTtoptextmargin}{2cm}
\newcommand{\CLASSINPUTbottomtextmargin}{2cm}
\newcommand{\CLASSINPUToutersidemargin}{1.9cm}
\newcommand{\CLASSINPUTinnersidemargin}{1.9cm}

\documentclass[10pt,a4paper]{IEEEtran}
\IEEEoverridecommandlockouts
\usepackage{cite}
\usepackage{amsmath,amssymb,amsfonts}
\usepackage{graphicx}
\usepackage{textcomp}
\usepackage{xcolor}
\usepackage{amsmath}
\usepackage{amssymb}
\usepackage{cite}
\usepackage{algpseudocode,algorithm,algorithmicx}
\usepackage{multirow}
\usepackage{url}
\usepackage{fancyhdr}
\usepackage[pdfencoding=auto, colorlinks=true, citecolor=black, filecolor=black, linkcolor=black, urlcolor=cyan, bookmarks=true, bookmarksopen=true, bookmarksnumbered=true, pdffitwindow=true, pdfpagemode=UseNone]{hyperref}
\usepackage{atbegshi}% http://ctan.org/pkg/atbegshi
\AtBeginDocument{\AtBeginShipoutNext{\AtBeginShipoutDiscard}}

\fancypagestyle{firstpage}{%
\vspace{1.5ex}
  \rhead{DRAFT}
}

\begin{document}
%\pagenumbering{gobble}

\title{\Huge On the role of ML estimation and Bregman divergences in sparse representation of covariance and precision matrices
\vspace{0.5ex}
}
\IEEEaftertitletext{\vspace{-1.5\baselineskip}}
\author{ \IEEEauthorblockN{Branko Brklja\v{c} and \v{Z}eljen Trpovski}}%,
\thanks{%
\vspace{-1.5ex}
\par{}This work was partially supported by the Ministry of Education, Science and Technological Development of the Republic of Serbia, as part of the project III44003.
\smallskip
\par{\IEEEauthorblockA{Branko Brklja\v{c} and \v{Z}eljen Trpovski are affiliated with the University of Novi Sad, Faculty of Technical Sciences, Department of Power, Electronic and Telecommunication Engineering, address: \textit{Trg Dositeja Obradovi\'{c}a~6, 21000 Novi Sad, Serbia}, (phone: +381~21~4852000, \mbox{e-mail}: \textit{brkljacb@uns.ac.rs}, \textit{zeljen@uns.ac.rs}).}}
}
%\\[3.0ex]
\maketitle
\pagestyle{empty}
\thispagestyle{firstpage}
\pagestyle{plain}
%\begin{abstract}
\hfill\break
\small\textbf{\indent \textit{Abstract} --- Sparse representation of structured signals requires modelling strategies that maintain specific signal properties, in addition to preserving original information content and achieving simpler signal representation. Therefore, the major design challenge is to introduce adequate problem formulations and offer solutions that will efficiently lead to desired representations. In this context, sparse representation of covariance and precision matrices, which appear as feature descriptors or mixture model parameters, respectively, will be in the main focus of this paper.}
%\end{abstract}

\setcounter{page}{1}

\normalsize
%\vspace{-2.2ex}
\section{Introduction}
%\vspace{-0.8ex}
%\vspace{-0.8ex}
\IEEEPARstart{W}{hen} present as a property, signal sparsity  proves to be very useful characteristic that brings many advantages in processing, analysis and inference. Second order tensors in the form of covariance and precision matrices, which can be regarded as structured signals, are also subject of sparse representation or some other, alternative modeling strategy. All these different approximation approaches aim towards achieving more efficient processing, estimation, or some other performance gains that are often possible to accomplish by using adequately constructed approximations of original signals (matrices). However, due to the structured nature of original signals, designed approximations usually need to preserve some additional properties of the signal that are detrimental for its role in some larger model or a system, in addition to signal's original information content. Therefore, there is a need to include constraints in the initial formulation of the corresponding approximation problem and offer solutions that will be efficient, but which will also preserve important signal properties. In addition, required approximation quality is usually determined by the user or application requirements, and can be a matter of specific choice, but original signal properties that are regarded as significant for characterizing certain family of signals always need to be fully preserved in the approximation process. In this paper we will limit the scope of our discussion to the field of pattern recognition and consider techniques for sparse approximation of covariance matrices and their inverses, which are better known as precision matrices.
\subsection{Covariance and precision matrices}
\label{subSection1.1}
\par{}Covariance matrix describes interactions between random variables (components of some random, feature vectors) and  represents discrete generalization of the covariance function, i.e. the second order joint central moment. Elements of the matrix can be regarded as the measures of linear dependence between corresponding pairs of features, while the matrix properties that are of particular interest are symmetry and positive semi-definiteness. %
\par{}However, there are different roles in which covariance matrices appear in signal processing and pattern recognition. Although they can be used to define whitening transforms, which decorrelate signal components, perform principal component analysis, or design optimal linear predictors in terms of minimal mean squared error (MSE), their roles that are of particular interest for the topic of this paper are those in which: a) covariances are used as agregate feature descriptors, or b) as parameters of individual mixture components in Gaussian mixture models (GMMs). %
\par{}In the latter case, each covariance defines the shape of the corresponding Gaussian, which has the functional form of the multivariate normal distribution. More exactly, inverse of each covariance, i.e. precision matrix, defines one of the corresponding quadratic forms that are computed during evaluation of the particular, class dependent, likelihood function of the statistical classifier. %
\par{}On the other hand, when used in the role of feature descriptor, each covariance represents a set of second order statistics that are collected over some local, e.g. spatial or temporal window or region, and which (as a set of specific values) comprehend linear dependences of feature vector components that are captured by multiple feature vector observations over such, predefined domain.
%
%
%
%\vspace{-2ex}
\subsection{Sparse approximation (representation)}
\label{subSection1.2}
\par{}Sparse approximations of covariance and precision matrices are usually designed in the form of sparse linear combinations of matrices that have the same dimension as originals, but which are chosen from some predefined or specially constructed set. As typical, term sparse means that only a relatively small number of set elements, nonzero terms in linear combination, is used for reconstruction of original signal. Depending on the type of approximation technique that is used, matrices in the previously mentioned set can be forced to have different characteristic properties, like the matrix symmetry only, which was imposed as a constraint in the case of sparse approximation of precision matrices utilized in GMMs that was proposed in \cite{brkljavc2014sparse}; or both symmetry and positive semi-definiteness, which were used in the case of the proposed sparse coding of image covariance descriptors in \cite{sivalingam2014tensor}.
\par{}Regardless of the type of the role that original symmetric, positive semi-definite matrices have in some recognition system (appearing as covariance feature descriptors, or as a mixture model parameters in the form of corresponding precisions), techniques for their sparse approximation, i.e. sparse representation, in both cases have a joint root in the fundamental way in which information distances based on Bregman matrix divergences enable formulations of the  initial  matrix approximation optimization problems. %
Namely, we will see that although motivations  from which  both mentioned sparse approximation techniques originate are very different, they share the same information preserving principles that result from the use of asymmetric Kullback-Leibler (KL) divergences between original and approximated probability distributions. More exactly, it will be shown that formulations of sparse representation of precision matrices in \cite{brkljavc2014sparse}, and sparse coding (representation) of image covariance descriptors in \cite{sivalingam2014tensor}, are both utilizing the same special case of Bregman matrix divergence corresponding to multivariate normal distributions, but offer solutions to different sparse representation problems due to the asymmetric nature of the corresponding divergence and different initial constraints, which are imposed on the members of sparse linear combinations that are used to reconstruct original precisions and covariances, respectively.%
%
%\vspace{-2ex}
\subsection{Aim of the paper}
\label{subSection1.3}
\par{}The aim of the paper is to give more insight into an interesting mutual relationship that exists between two different formulations of sparse representation problem, which were proposed in the previously mentioned studies \cite{sivalingam2014tensor} and \cite{brkljavc2014sparse}. It will be presented through a theoretical analysis of the role of maximum likelihood (ML) estimation of Gaussian mixture model parameters in sparse representation of precision matrices and its connection to information distances based on Bregman matrix divergences. In addition, although proposed algorithmic solutions in \cite{sivalingam2014tensor} and \cite{brkljavc2014sparse}  are: a)~based on different frameworks, generally speaking on quadratic programming in the first case, and an iterative shrinkage-thresholding algorithm with an active set strategy in the second case; and b)~address different types of problems, sparse representation of covariances and precisions; both methods try to achieve the same goal: sparse representation that will preserve original information content of positive definite matrices.
\par{}The rest of the paper is organized as follows: in Section~\ref{Section2} an asymptotic equivalence between maximum likelihood estimation and the principle of minimum discriminative information is discussed, then in Section~\ref{Section3} a family of information distances based on Bregman matrix divergences is introduced, which is followed in Section~\ref{Section4} by discussion of two different sparse representation formulations that are used in sparse approximation of covariance and precision matrices, and which respectively, appear either as feature descriptors or mixture model parameters in some recognition system. Finally, in Section~\ref{Section5} we give some concluding remarks and point out future research directions.
\section{ML estimation and the principle of minimum discrimination information}
\label{Section2}
%\vspace{-2ex}
\subsection{Optimality of ML estimation}
\label{subSection2.1}
\par{}Generally speaking, maximum likelihood estimation of unknown random parameters in some model of interest can be considered as optimal in the Bayesian sense in the case when the chosen optimality criterion is minimum expected MSE of obtained estimates and when no apriori information about distribution of parameter values is available. More exactly, ML can be considered as optimal under given circumstances (quadratic cost and uninformative, uniform prior) if the corresponding posterior distribution of unknown parameter values is assumed to be Gaussian. In such case, ML optimality arises from the symmetric nature of the corresponding likelihood function of unknown parameters, which is under adopted assumption also Gaussian. In addition, described optimality of ML in the Bayesian sense also extends to other cost function choices, such as absolute or zero-one error, since the mean, median and mode of Gaussian distribution always coincide. %
\par{}Namely, if we denote with $C(\hat{\boldsymbol{\tau}}, \boldsymbol{\tau}) =\|\hat{\boldsymbol{\tau}}-{\boldsymbol{\tau}}\|^2_2$, quadratic cost function of an estimate $\hat{\boldsymbol{\tau}}$, which based on observations $\mathbf{x}$ minimizes the average estimation cost:
\begin{equation}\label{Rizik_estimacije}
\mathcal{R} (\hat{\boldsymbol{\tau}}) =\int\limits_{\{\boldsymbol{\tau}\}}\int\limits_{\{\mathbf{x}\}}{
C(\hat{\boldsymbol{\tau}}, \boldsymbol{\tau}) \, p(\mathbf{x}, \boldsymbol{\tau})}\,{\rm {d}}\mathbf{x} \, {\rm {d}}\boldsymbol{\tau}\,,
\end{equation}
then after differentiation with respect to $\hat{\boldsymbol{\tau}}$, from the condition of optimality $\partial \mathcal{R}/ \partial \hat{\boldsymbol{\tau}} = 0$, it follows that the minimum expected MSE in~\eqref{Rizik_estimacije} is determined by the solution of:
\begin{equation}\label{Rizik_estimacije_2}
\int\limits_{\{\mathbf{x}\}}
\;
\Bigl\{\,
{
\int\limits_{\{\boldsymbol{\tau}\}}
{
(\hat{\boldsymbol{\tau}}-{\boldsymbol{\tau}}) \, p(\boldsymbol{\tau}|\mathbf{x})
\, {\rm {d}}\boldsymbol{\tau}
}
}\,
\Bigr\}
\;
{p(\mathbf{x}) \,{\rm {d}}\mathbf{x} }
\,=\,0  \,,
\end{equation}
which gives optimal Bayesian estimate of unknown parameters in the form of conditional expectation of $\boldsymbol{\tau}$:
\begin{equation}\label{Rizik_estimacije_3}
\hat{\boldsymbol{\tau}}=
\mathbb{E}[\left. \boldsymbol{\tau} \right|\mathbf{x}]  =\int\limits_{\{\boldsymbol{\tau}\}}
{\boldsymbol{\tau} \, p(\boldsymbol{\tau} |\mathbf{x})}\,{\rm {d}}\boldsymbol{\tau}
\,.
\end{equation}
Risk in \eqref{Rizik_estimacije} was deliberately named as expected MSE in order to point out that unknown parameters are considered as random quantities, and that MSE of their estimate is averaged over different observations.
In the case when $p(\boldsymbol{\tau} |\mathbf{x})$ is assumed to be Gaussian and $p(\boldsymbol{\tau})$ uniform, since mean and mode of distribution are the same, $\hat{\boldsymbol{\tau}}$ in \eqref{Rizik_estimacije_3} becomes equal to the corresponding ML estimate of unknown parameters:
\begin{equation}\label{Rizik_estimacije_4}
\hat{\boldsymbol{\tau}}=
{\rm{mode}}( p(\boldsymbol{\tau} |\mathbf{x})) = \underset{\boldsymbol{\tau}}{\mathop{\max }}\, \frac{{p(\mathbf{x}|\boldsymbol{\tau}) \, p(\boldsymbol{\tau})}}{{p(\mathbf{x})}} = \underset{\boldsymbol{\tau}}{\mathop{\max }}\, p(\mathbf{x}|\boldsymbol{\tau})\,.
\end{equation}
Gaussian assumption about posterior $p(\boldsymbol{\tau} |\mathbf{x})$ also implies that the structure of conditional expectation in \eqref{Rizik_estimacije_3} will be linear if the adopted prior $p(\boldsymbol{\tau})$ is also Gaussian, \cite{kay1993fundamentals}. %
\par{}However, more important is that regardless of the shape of  $p(\boldsymbol{\tau} |\mathbf{x}) $, i.e. when Gaussian assumption is discarded, maximum aposteriori (MAP) estimate in \eqref{Rizik_estimacije_4} is always optimal in Bayesian sense, in terms of zero-one cost function, which means that under adopted assumption of an uninformative prior $p(\boldsymbol{\tau})$ (or when the prior can be regarded as uniform over the domain where posterior is essentially nonzero), ML estimate can be also considered as optimal in Bayesian sense, in such case, \cite{whalen1971detection}. Moreover, in the classical setting, where unknown parameter values are presumed to be deterministic, ML estimation is usually considered as asymptotically unbiased and efficient in the presence of large data samples. In addition, it is also considered as optimal in the finite sample case of such setting, if the relationship between noisy observations and parameters is assumed to be linear, and the nature of noise regarded as Gaussian. As a special case, if the noise samples are also uncorrelated and with the same variance, ML estimate is equivalent to the least squares solution, which in that specific case results in an efficient estimate.%
\subsection{Minimum discrimination information}
\label{subSection2.2}
\par{}Principle of minimum discriminative information was proposed in an effort to offer unification of statistical hypothesis testing approaches through consistent application of information theory concepts,\cite{kullback1968information}. Decision or recognition process should be viewed as a problem of discrimination based on measurement of minimal distance or divergence between statistical populations, and it should be based on quantity that describes information distance between two distributions. After being introduced in \cite{kullback1951information}, this information measure became widely known as Kullback-Leibler (KL) divergence, or relative entropy, \cite{cover1991elements}:
\begin{equation}\label{MDI_1}
D_{\mathrm {KL} }(P\|Q)\triangleq \int _{-\infty }^{\infty }p(x)\,\log {\frac {p(x)}{q(x)}}\,{\rm {d}}x \,,
\end{equation}
where $p(x)$ and $q(x)$ are probability density functions of distributions $P$ and $Q$, and \eqref{MDI_1} represents directed divergence from $Q$ to $P$. Interpretation of this nonnegative quantity depends on the context of corresponding distributions, however it is always directed, and there are always "first" and "second" distribution, $p$ and $q$, respectively. Originally, \cite{kullback1951information}, it was denoted as $I_{1:2}(E)$ or $I(1:2)$ and interpreted as generalization of logarithmic information measure proposed in \cite{shannon1948}, in the case of continuous distributions, while under term divergence was considered its symmetrized version in the form of a sum: $I_{1:2}(X)+I_{2:1}(X)$.
\par{}Principle of minimum discrimination information (MDI) states that the right decision comes from the best match of the shape of estimated parameterized distribution with one of the shapes of learned or "true" distributions, which were associated in advance with each of the possible hypothesis or categories. Best match corresponds to the minimum difference in information contained in estimated and one of the learned distributions, as measured by the directed divergence \eqref{MDI_1}. It was shown, \cite{kullback1968information}, that such reasoning strategy leads to the same decision rules that arise from the special cases of Bayesian hypothesis testing, e.g. presented in \cite{helstrom1975statistical} or \cite{whalen1984statistical}. MDI also demonstrates the significance of the design of discriminative features for the overall performance of the recognition system, i.e. importance of maximization of differences between samples of different categories. In the context of information transfer, MDI also enables design of the code with expected length of message that tends to source entropy $H(p)$, which is equivalent to minimization of the corresponding cross-entropy between true distribution $p$ and estimate $q$:
\begin{equation}\label{MDI_2}
H(p,q)=\mathbb {E} _{p}[-\log q]=H(p)+D_{\mathrm {KL} }(p\|q).
\end{equation}
\par{}Therefore it should be said that for achieving minimum risk reasoning in some decision system, in addition to having good estimation of corresponding category distributions, it is essential to design features that will describe such categories in the most discriminative way. In the context of estimation, \cite{lovric2011international}, which is the most important from the perspective of this paper, KL divergence has always had a significant role as a basis for the design of various information criterions, \cite{akaike1974new}. More exactly, criterion in \cite{akaike1974new} was one of the first examples in which formulation based on \eqref{MDI_1} or \eqref{MDI_2} was extended with a regularization term that had taken into account model's complexity. Original optimization objective based on MDI is extended with the aim to avoid overfitting, but also in order to achieve additional goals, like model selection. Similar strategy is also employed in the sparse representation of covariance and precision matrices in \cite{sivalingam2014tensor} and  \cite{brkljavc2014sparse}.
\subsection{ML estimation as KL divergence minimization}
\label{subSection2.3}
Finally, it will be shown that parameter estimation based on the ML criterion asymptotically converges to minimization of corresponding KL divergence, or that it is asymptotically equivalent to minimization of minimum discrimination information between parameterized distribution based on ML estimates (distribution approximation) and parameterized distribution based on the true parameter values.
Suppose that there are $n$ i.i.d. observations, $\mathcal{X} = \{x_i\}_{i=1}^n$, generated by the true distribution $p_{\theta^*}(x)$, then their joint distribution can be factored into: $p_{\theta}(x_1,x_2, \ldots, x_n ) = p({x_1|\theta}) p({x_2|\theta})\cdots p({x_
n|\theta})$, where $p_{\theta^*}(x_i) = p({x_i|\theta^*})$ are true marginal distributions, while the corresponding ML estimate of unknown parameter $\theta \in \Theta$ is given by:
\begin{equation}\label{ML_MDI_1}
\hat{\theta} = \underset{\theta \in \Theta}{\mathop{\arg \max }}\:{ \prod_{i=1}^{n} p_{\theta}(x_i)  }\,.
\end{equation}
\par{}Objective in \eqref{ML_MDI_1} depends on the modeling approach used for description of $p({x_i|\theta})$. However, optimality of estimate \eqref{ML_MDI_1} with respect to true $\theta^*$ is determined by conditions expressed in Section \ref{subSection2.1}, regardless of the quality of the solution $\hat{\theta}$ that is obtained by the corresponding optimization procedure that is utilized in \eqref{ML_MDI_1}, which is certainly important by itself. Nevertheless, formulation presented in \eqref{ML_MDI_1} appears very often and proves to be very useful if $p_{\theta}(x_i)$ have exponential form, since the monotonically increasing logarithmic function enables one to replace the product in \eqref{ML_MDI_1} with the sum of corresponding log-likelihood functions:
\begin{equation}\label{ML_MDI_2}
\hat{\theta} = \underset{\theta \in \Theta}{\mathop{\arg \max }}\:{ \sum_{i=1}^{n} \ln p_{\theta}(x_i)  }\,,
\end{equation}
which also can be rewritten as minimization of the sum of negative log-likelihoods:
\begin{equation}\label{ML_MDI_3}
\begin{split}
\hat{\theta} = \underset{\theta \in \Theta}{\mathop{\arg \min }}\: {\mathcal{L(\theta; \mathcal{X})}}
=\underset{\theta \in \Theta}{\mathop{\arg \min }}\:{\frac{1}{n} \sum_{i=1}^{n} - \ln p_{\theta}(x_i)  } \,.
\end{split}
\end{equation}
Objective ${\mathcal{L(\theta; \mathcal{X})}}$ in \eqref{ML_MDI_3} incorporates a scaling factor $1/n$ in order to better resemble the sample mean, which almost surely tends towards expected value in the large sample case, $n \to \infty$. More exactly:
\begin{equation}\label{ML_MDI_4}
P\left( \;\lim_{ n \to \infty}{\bar{x}_n}   = \mu \;\right)  = 1\, \Leftrightarrow \, {\bar{x}_n} \xrightarrow[n \to \infty]{\text{~a.s.~}} \,\mu=\mathbb{E}[x_i] \, .
\end{equation}
In the case of  $n \to \infty$, when applied to ${\mathcal{L(\theta; \mathcal{X})}}$ from \eqref{ML_MDI_3}, presented limit in \eqref{ML_MDI_4} results in the following identity:
\begin{equation}\label{ML_MDI_5}
{\frac{1}{n} \sum_{i=1}^{n} - \ln p_{\theta}(x_i)  } \xrightarrow[]{\text{~a.s.~}} \, \mathbb{E}_{p_{\theta^*}}\left[-\ln p_{\theta}(x) \right] \,,
\end{equation}
where expectation is performed over distribution $p_{\theta^*}$, which is parameterized by the unknown true parameter values $\theta^*$. Objective ${\mathcal{L(\theta; \mathcal{X})}}$ in \eqref{ML_MDI_3} could achieve minimum for the ideally chosen parameters $\theta^*$, which have generated observed data. Therefore, based on \eqref{ML_MDI_5},  its  asymptotic value would be given by the value of: $\mathbb{E}_{p_{\theta^*}}\!\left[-\ln p_{\theta^*}(X) \right]$. Since this value would be less or equal than any other objective value, obtained for any other choice of $\theta \in \{\Theta\setminus{\theta^*}\}$, it follows that the objective ${\mathcal{L(\theta; \mathcal{X})}}$  in \eqref{ML_MDI_3} can be replaced by the following minimization problem, where the aim is to minimize the difference:
\begin{equation}\label{ML_MDI_6}
\Delta = \mathbb{E}_{p_{\theta^*}}\!\left[-\ln p_{\theta}(x) \right] - \mathbb{E}_{p_{\theta^*}}\!\left[-\ln p_{\theta^*}(x) \right] \,,
\end{equation}
which, based on the linearity of expectation operator, gives:
\begin{equation}\label{ML_MDI_7}
\begin{split}
\Delta & =  \mathbb{E}\left[\,\ln p_{\theta^*}(x) \,-\,\ln p_{\theta}(x) \,\right] =
\mathbb{E}\left[\, \ln \frac{ p_{\theta^*}(x)}{ p_{\theta}(x)} \,\right] \\
& = \int _{-\infty }^{\infty }p_{\theta^*}(x)\,\ln {\frac{\ p_{\theta^*}(x)}{ p_{\theta}(x)}}\,{\rm {d}}x\,.
\end{split}
\end{equation}
\par{}The last expression in \eqref{ML_MDI_7} represents the KL divergence, introduced in \eqref{MDI_1}, Section \ref{subSection2.2}, which measures difference in information content between: true or original distribution $p_{\theta^*}$, and its approximation $p_{\theta}$, which is based on some current estimate of paramaters, $\theta$, computed by the utilized optimization procedure that will hopefully lead to some final estimate $\hat{\theta}$. %
\par{}This means that the presented ML estimation procedure in the asymptotic case transforms itself into corresponding MDI principle, which tries to reduce $D_{\mathrm {KL} }(p_{\theta^*}\|p_{\theta})$. In order to prove that (besides their asymptotic equivalence in the large sample case) both estimation approaches also converge towards the true parameter values $\theta^*$, consider the sample estimate $\bar{\Delta}$ of the previously defined expected difference~$\Delta$ in \eqref{ML_MDI_6} and \eqref{ML_MDI_7}. Namely, just for a moment suppose that ${\hat{\theta}_n}$ is the best solution of \eqref{ML_MDI_3} in the case of $n$ samples, i.e. one that minimizes ${\mathcal{L(\theta; \mathcal{X})}}$ even better than the true $\theta^*$, then the $\bar{\Delta}$ of the corresponding negative log-likelihoods would be nonpositive and given by:
\begin{equation}\label{ML_MDI_8}
\begin{aligned}
\bar{\Delta}& \stackrel{\text{\eqref{ML_MDI_6}}}{=} \,
{\mathcal{L}({\hat{\theta}_n}; \mathcal{X})} - {\mathcal{L}({\theta^*}; \mathcal{X})} \\
&\, \stackrel{\text{\eqref{ML_MDI_3}}}{=}
\,  \frac{1}{n}\sum_{i=1}^n{\ln p_{\theta^*}(x_i)}
 -  \frac{1}{n} \sum_{i=1}^n{ \ln p_{\hat{\theta}_n}(x_i)}
\; \leq \,0
\, ,
\end{aligned}
\end{equation}
however, it will be shown that such (wrong) assumption would be possible only if ${\hat{\theta}_n}$ in the limit becomes $\theta^*$. Given that $D_{\mathrm {KL} }(p\|q) $  is always $\geq 0$, inequality in \eqref{ML_MDI_8} can be transformed into:
\begin{equation}\label{ML_MDI_9}
\begin{aligned}
&
{\frac{1}{n}}
\sum_{i=1}^n{\ln \frac{ p_{\theta^*}(x_i)}{ p_{\hat{\theta}_n}(x_i)}}
+ D_{\mathrm {KL} }(p_{\theta^*}\|p_{\hat{\theta}_n})  \leq
 D_{\mathrm {KL} }(p_{\theta^*}\|p_{\hat{\theta}_n})
 \\
&\Leftrightarrow
D_{\mathrm {KL} }(p_{\theta^*}\|p_{\hat{\theta}_n})  \leq
 \left|{D_{\mathrm {KL} }(p_{\theta^*}\|p_{\hat{\theta}_n})
 -
\frac{1}{n} \sum_{i=1}^n{\ln \frac{ p_{\theta^*}(x_i)}{ p_{\hat{\theta}_n}(x_i)}}} \right|
\,.
\end{aligned}
\end{equation}
Taking into account previous considerations, presented in \eqref{ML_MDI_5}, it follows that the average sum of logaritmic ratios in \eqref{ML_MDI_9},  in the limit when $n \to \infty$, should almost surely be equal to the corresponding expected value that is, by definition \eqref{MDI_1} since $p_{\theta^*}$ is the data generating distribution, given by the KL divergence $D_{\mathrm {KL} }(p_{\theta^*}\|p_{\hat{\theta}_n}) $. This is exactly the second term on the right-hand side of the last inequlity in \eqref{ML_MDI_9}, which results in the following assertion:
\begin{equation}\label{ML_MDI_10}
\left|{\frac{1}{n}\,\sum_{i=1}^n{\ln \frac{ p_{\theta^*}(x_i)}{ p_{\hat{\theta}_n}(x_i)}} \, - \,D_{\mathrm {KL} }(p_{\theta^*}\|p_{\hat{\theta}_n})}\right| \, \xrightarrow[n \to \infty]{\text{~a.s.~}} \, 0
\,,
\end{equation}
Since $D_{\mathrm {KL} }$ is always nonnegative, in the limit case when $n \to \infty$, $D_{\mathrm {KL} }(p_{\theta^*}\|p_{\hat{\theta}_n})$  in \eqref{ML_MDI_9}   must become zero due to the previous result in \eqref{ML_MDI_10},  which forces the right hand side of the of the last inequality in \eqref{ML_MDI_9} towards zero.
According to MDI principle, this is equivalent to $p_{\hat{\theta}_n}\!  \xrightarrow[]{} \! p_{\theta^*}$, which means that ML estimate $\hat{\theta}_n$, in the limit case when $n \to \infty$, also converges towards the same true parameter values as MDI procedure, i.e. $\hat{\theta}_n \xrightarrow[]{} \theta^*$.
\section{Bregman matrix divergences as information distances}
\label{Section3}
In Section \ref{Section2} we have already seen an important class of directed information distances. However, when considering design of sparse representation formulation in the case of structured signals such are covariance and precision matrices, it is useful to have complementary perspective on the corresponding sparse approximation problem in which we are essentially interested in estimation of unknown sparse representation parameters, Section \ref{Section4}. Therefore, some emphasis in this paper will be put on the rich interaction that exists between a special class of KL divergences corresponding to multivariate normal distributions, on one side, and a family of information distances based on Bregman matrix divergences (BMD) between positive semi-definite matrices, on the other. Although the latter group is much broader and appears in some novel applications, most of the discussion will be oriented towards KL divergences between normal distributions and the role of  covariance matrices and precisions, which parameterize normal distributions and GMMs .
\subsection{Vector induced BMD}
\label{subSection3.2}
\par{}If $\varphi:\mathbb{R}^d \rightarrow \mathbb{R}$, is differentiable function over convex domain, Bregman vector divergence $D_\varphi$ is defined as approximation error $\varphi(x)$ that results from the use of the first order Taylor expansion of $\varphi$ in the vincinity of $y$:
\begin{equation}\label{B_1}
D_{\varphi}(x,y) = \varphi(x) - \big(\;\varphi(y) \,+ \,(x-y)^T\,\nabla\varphi(y)\;\big)\,.
\end{equation}
Introducing matrix notation, if the scalar product between matrices $\rm{A}$ and $\rm{B}$ is defined as: $\langle {\rm{A}}, {\rm{B}} \rangle = \mathrm{tr} \left({{\rm{A}} }^{T}{{\rm{B}} }\right)$, and $\nabla\varphi({{\rm{A}} })$ denotes a gradient of the scalar function of matrix $A$, Bregman matrix divergence (BMD) is defined in a similar manner as in \eqref{B_1}, by:
\begin{equation}\label{B_2}
D_{\varphi}({{\rm{A}} },{{\rm{B}} }) = \varphi({{\rm{A}} }) - \varphi({{\rm{B}} }) - \mathrm{tr} \big(\left({{\rm{A}} }-{{\rm{B}} }\right)^T \nabla\varphi({{\rm{B}} }) \big) \,.
\end{equation}
Different divergence types are determined by the choice of scalar function ${\varphi}$, e.g. Frobenius matrix norm, $\varphi(\cdot) = \|\cdot \|^2_F$, results in: $D_{\mathrm {F} }({{\rm{A}} },{{\rm{B}} })  = \| {{\rm{A}} } - {{\rm{B}} }\|^2_F$, which is just a matrix generalization of a vector Bregman divergence induced by the square of Euclidean norm, $\|\cdot \|^2_2$. Such analogy between vector and matrix setting is not an exception, as will be demonstrated by the next example. Consider a function $\varphi(x) = \sum\nolimits_i\left(x_i \ln x_i - x_i \right)$, where $x = (x_1,...,x_d)^T$. According to \eqref{B_1}, after replacing gradient components: $\partial \varphi(y)/\partial y_i = \ln y_i$, in \eqref{B_1}, we get the following divergence between points $x$ and $y$ in $\mathbb{R}^d$:
\begin{equation}\label{B_3}
D_{\varphi}(x, y) =  \sum\nolimits_{i=1}^d \Big(x_i \ln \frac{x_i}{y_i} - x_i + y_i \Big) \,.
\end{equation}
Closer look at \eqref{B_3} reveals that if there would exist an additional normalization of corresponding vectors, which would force them to satisfy the condition: $\sum\nolimits_{i=1}^d \tilde{x}_i = 1$, i.e. make $\tilde{x}$ and $\tilde{y}$ lying on simplex in $\mathbb{R}^d$, convex function $\varphi(\tilde{x})$ would represent a negative entropy of probability mass function defined by $\tilde{x}$. This implies that after described normalization, $D_{\varphi}$ in  \eqref{B_3} actually represents a KL divergence: $D_{\mathrm {KL} }(\tilde{x}\,\|\,\tilde{y})$, eq. \eqref{MDI_1}, between discrete distributions $\tilde{x}$ and $\tilde{y}$, which proves that \eqref{MDI_1} in essence is only a type of Bregman (matrix) divergences, which is generated by the negative of entropy function from \cite{shannon1948}.
\par{}Following the same route as in the case of \eqref{B_3}, by defining the scalar function which corresponds to nonnormalized entropy function of matrix eigenvalues:
\begin{equation}\label{B_4}
\varphi({{\rm{A}} }) = \sum\nolimits_{i=1}^d\left(\lambda_i \ln \lambda_i - \lambda_i \right) \,,
\end{equation}
where $\lambda_i$ are corresponding eigenvalues of symmetric matrix ${{\rm{A}} }$, defined by the similarity transformation:  ${{\rm{\Lambda}} } = {{\rm{Q}}}^T {{\rm{A}}}{{\rm{Q}}}$, we will arrive to a family of analogous BMD. Since $\mathrm{tr} \left( {{\rm{A}} } \right) = \mathrm{tr} \left( {{\rm{Q}}}{{\rm{\Lambda}}}{{\rm{Q}}}^T \right) =  \mathrm{tr} \left( {{\rm{\Lambda}}}{{\rm{Q}}}{{\rm{Q}}}^T \right) = \mathrm{tr} \left( {{\rm{\Lambda}}} \right)= \sum\nolimits_i \lambda_i$, and if matrix logarithm is defined as: $\ln{{\rm{A}} } = {{\rm{Q}}}\ln{{\rm{\Lambda}}}{{\rm{Q}}}^T$, where  $\ln{{\rm{\Lambda}}}$ denotes diagonal matrix of logarithmic eigenvalues, original  $\varphi({{\rm{A}} })$ can be rewritten as:  %
\begin{equation}\label{B_5}
\varphi({{\rm{A}} }) = \mathrm{tr} \big({{\rm{A}} } \ln{{\rm{A}} } - {{\rm{A}}}\big)  \,,
\end{equation}
since: $\mathrm{tr} \big({{\rm{A}}}\ln{{\rm{A}} }-{{\rm{A}}}\big) =
\mathrm{tr}\big( {{\rm{Q}}}\big({{\rm{\Lambda}}} \ln {{\rm{\Lambda}}} - {{\rm{\Lambda}}} \big){{\rm{Q}}}^T\big) = {\textrm{eq.~}}\eqref{B_4}$, %
which after replacing \eqref{B_5} in \eqref{B_2}, results in:
\begin{equation}\label{B_6}
D_{\mathrm {N} }\big({{\rm{A}} },{{\rm{B}} }\big)= \mathrm{tr} \big( {{\rm{A}}} \ln{{\rm{A}} } - {{\rm{A}}} \ln{{\rm{B}} } - {{\rm{A}}} + {{\rm{B}}}\big)\,,
\end{equation}
the so called Neumann BMD, or quantum relative entropy.
\par{}However, in the context of sparse representation formulation considered in this paper, we are far more interested in the next example of BMD, which is induced by the sum of logarithms of matrix eigenvalues, i.e. which is related to matrix determinant or "volume".
\subsection{LogDet divergence}
\label{subSection3.2}

Namely, since $\det {{\rm{A}} } = \prod\nolimits_{i=1}^d\lambda_i$, if corresponding scalar function is defined to be:
\begin{equation}\label{B_7}
\varphi({{\rm{A}} }) = \sum\nolimits_{i=1}^d - \ln \lambda_i = - \ln \det {{\rm{A}} } = - \mathrm{tr}  \ln {{\rm{A}} }
\,,
\end{equation}
which is also known as Burg entropy of matrix eigenvalues, resulting BMD is LogDet or Burg directed divergence:
\begin{equation}\label{B_8}
D_{\mathrm {B} }\big({{\rm{A}} },{{\rm{B}} }\big) =
 \mathrm{tr} \big({{\rm{A}}} {{\rm{B}}}^{-1} \big) - \ln \det \big( {{\rm{A}}} {{\rm{B}}}^{-1}  \big)  - d\,.
\end{equation}
Terms in which matrices ${{\rm{A}} } $ and ${{\rm{B}} } $ appear in \eqref{B_8} are always in the form of product ${{\rm{A}}} {{\rm{B}}}^{-1}$. Since ${{\rm{A}} } $ and ${{\rm{B}} } $ can be considered to be either covariances or precisions, their inversions are also positive definite and it can be shown that eigenvalues of  product  ${{\rm{A}}} {{\rm{B}}}^{-1}$ will be also positive, which means that \eqref{B_8} can be fully characterized by them.
If ${{\rm{A}} } $ and ${{\rm{B}} } $ are decomposed on corresponding sets of orthonormal eigenvectors $\{q_i\}$ and $\{v_i\}$, and eigenvalues $\{\lambda_i\}$ and $\{\psi_i\}$,  respectively, which are given by ${{\rm{A}} }  = {{\rm{Q}}}{{\rm{\Lambda}}}{{\rm{Q}}}^T$ and ${{\rm{B}} } = {{\rm{V}}}{{\rm{\Psi}}}{{\rm{V}}}^T$, LogDet divergence can be equivalently rewritten as:
\begin{equation}\label{B9}
D_{\mathrm {B} }\left({{\rm{A}} },{{\rm{B}} }\right)=  \sum_{i=1}^d \sum_{i=1}^d \frac{\lambda_i}{\psi_j}\left( q_i^T v_j \right)^2 - \sum_{i=1}^d \ln \frac{\lambda_i}{\psi_i} - d\,.
\end{equation}
Presented formulation of LogDet in \eqref{B9} shows that directed distance depends on the similarity of corresponding pairs of normalized matrix eigenvectors, which is equivalent to measuring angle between them, and also on the similarity of corresponding pairs of eigenvalues. It means that in the case of ${{\rm{A}} } = {{\rm{B}} } $, double sum in \eqref{B9} would collapse into sum of $d$ ones, while the second sum would become zero. Similarly to previously described symmetrization approach, which was used in \cite{kullback1951information}, Section~\ref{subSection2.2}, directed Burg divergence in \eqref{B_8} can be also symmetrized as:
\begin{equation}\label{B10}
\begin{aligned}
J\left({\rm{A}},{\rm{B}}\right) &= {1 / 2}\;  D_{\mathrm {B} }\left({{\rm{A}} },{{\rm{B}} }\right)  + {1 / 2}\;  D_{\mathrm {B} }\left({{\rm{B}} }, {{\rm{A}} }\right)\\
&= {1 / 2}\;\mathrm{tr} \left( {{\rm{A}} }{{\rm{B}} }^{-1}  \right) + {1 / 2}\;\mathrm{tr} \left({{\rm{B}} }{{\rm{A}} }^{-1}  \right)  - d\,,
\end{aligned}
\end{equation}
which is known as Jeffreys matrix divergence. However, symmetric property does not immediately mean that it is a metric. Alternative symmetrization of \eqref{B_8} which was recently proven to be a metric is Stein or S-divergence,~\cite{sra2016positive}.
\subsection{LogDet and KL divergence}
\label{subSection3.2}
Between directed Burg divergence \eqref{B_8} of two positive definite matrices, on one side, and KL divergence~\eqref{MDI_1} of two Gaussian distributions, on the other, exists direct equivalence under precondition that mean values of Gaussian distributions that are compared using either directed distance are the same. Therefore it can be concluded that although LogDet is natural Bregman matrix divergence, it actually represents a special case of KL divergence of normal distributions, i.e. a special case of directed information distance. Therefore, it is suitable to be used as an objective measure of the quality of constructed sparse approximations of signals such are covariance and precision matrices, i.e. it corresponds to preserved information content of original structured signals. However, in order to achieve additional properties of designed approximations, such as sparsity, it must be combined with other objective terms, as will be briefly sketched in Section~\ref{Section5}. In the following lines some more details of this interesting relationship between LogDet and KL divergence will be presented.
\par{}Since covariances represent characteristics of the corresponding Gaussian distributions, difference in shapes of two Gaussian distributions, their information distance, can be described by the corresponding matrix divergence. This can be shown if we consider KL divergence between two multivariate normal distributions, of the functional form:
\begin{equation}\label{B11}
p(x)=\frac{1}{{{\left( 2\pi  \right)}^{d/2}}{{\left| \Sigma  \right|}^{1/2}}} \exp \left[-\frac{1}{2}
{{(x-\mu )}^{T}}{{\Sigma }^{-1}}(x-\mu ) \right]\,, %
\end{equation}
where $\Sigma$ is covariance, and ${\Sigma }^{-1}$ corresponding precision, while the mean vector is denoted with $\mu$. %
\par{}Starting from \eqref{MDI_1}, denoting corresponding covariance matrices of Gaussian distributions ${\mathcal {N}}_{{\rm{A}} }$ and ${\mathcal {N}}_{{\rm{B}} }$, as $\Sigma_{\rm{A}}$ and  $\Sigma_{\rm{B}}$, and by utilizing the following matrix identity:
\begin{equation}\label{B12}
\begin{aligned}
{\mathbb{E}}&_{p_{{\rm{A}} }} \bigl[  \left(x-\mu _{{\rm{B}} }\right)^{T}\Sigma _{{\rm{B}} }^{-1}(x-\mu _{{\rm{B}} })  \bigr] = \\
&= \left(\mu _{{\rm{A}} }-\mu _{{\rm{B}} }\right)^{T}\Sigma _{{\rm{B}} }^{-1}(\mu _{{\rm{A}} }-\mu _{{\rm{B}} }) + \mathrm {tr} \left(\Sigma _{{\rm{B}} }^{-1}\Sigma _{{\rm{A}} }\right)\,,
\end{aligned}
\end{equation}
corresponding KL divergence between "approximation" ${\mathcal {N}}_{{\rm{B}} }$ and "original" distribution ${\mathcal {N}}_{{\rm{A}} }$ can be defined as:
\begin{equation}\label{B13}
\begin{aligned}
&D_{\mathrm {KL} }\big({\mathcal {N}}_{{\rm{A}} }\|{\mathcal {N}}_{{\rm{B}} }\big)=
{1 \over 2}\Big[
\ln \left({\det \Sigma _{{\rm{B}} } \over \det \Sigma _{{\rm{A}} }}\right) -
d +
\mathrm {tr} \left(\Sigma _{{\rm{B}} }^{-1}\Sigma _{{\rm{A}} }\right)+ \\
&\;\;\;\;\;\;\;\;\;\;\;\;\;\;\;\;\;\;\;\;\;\;\;\;\;\;\;\;\;\;
+\left(\mu _{{\rm{B}} }-\mu _{{\rm{A}} }\right)^{T}\Sigma _{{\rm{B}} }^{-1}(\mu _{{\rm{B}} }-\mu _{{\rm{A}} })\Big]\,,
\end{aligned}
\end{equation}
where term $d$ results from: $d=\mathrm {tr} \left(\Sigma _{{\rm{A}} }^{-1}\Sigma _{{\rm{A}} }\right)$.
\par{}Symmetrized version of \eqref{B13}, which was mentioned in Section~\ref{subSection2.2} in the form of: $I_{1:2}(X)+I_{2:1}(X)$, can be obtained as an average of $D_{\mathrm {KL} }\big({\mathcal {N}}_{{\rm{A}} }\,\|\,{\mathcal {N}}_{{\rm{B}} }\big)$ and $D_{\mathrm {KL} }\big({\mathcal {N}}_{{\rm{B}} }\,\|\,{\mathcal {N}}_{{\rm{A}} }\big)$, which, if we assume that $\mu _{{\rm{A}} }$ is equal to $\mu _{\rm{B}}$, gives:
\begin{equation}\label{B14}
J_\mathrm {KL}\left({\rm{A}},{\rm{B}}\right) ={1 \over 2}\; {\mathrm{tr}}\, \Big(\,
\Sigma _{{\rm{A}} }^{-1}\Sigma _{\rm{B}} \,+\, \Sigma _{{\rm{B}} }^{-1}\Sigma _{{\rm{A}}} \,-\, 2\,{{\rm{I}}}\,\Big)
\,,
\end{equation}
where ${\rm{I} }$ denotes identity matrix of size $d$. %
\par{}The first thing to notice is that expression in \eqref{B14} is equivalent to \eqref{B10}, which represents a symmetrized version of LogDet divergence, since trace is linear and invariant to cyclic permutations of factors in a product, while the trace of diagonal matrix ${\rm{I} }$ reduces to $d$. However, instead of symmetrized versions, we are more interested in equivalence between original asymmetric divergences in \eqref{B13} and \eqref{B_8}.  %
\par{}This can be seen if we replace $\mu _{{\rm{A}} } = \mu _{\rm{B}}$ in \eqref{B13}, which removes the corresponding quadratic form from \eqref{B13}, and results in expression \eqref{B_8}, since product of determinants is equal to determinant of a product, while the reciprocal of determinant is equal to determinant of inverse (also taking into account that logarithm of determinant in  \eqref{B_8}  appears with the negative sign). %
\par{}Presented analysis offers an insight into similarities between these different formulations of the same information distance. However, as will be shown in the next Section, due to asymmetric nature of the presented information distance, if used as part of an optimization objective, it can result in two different types of optimization problems, depending on the "place" of original and approximated distribution, which can be regarded as arguments of expression \eqref{B13}, i.e. corresponding covariances or precisions in the equivalent case of expression \eqref{B_8}.
\section{Sparse representation of covariance and precision matrices}
\label{Section4}
Under sparse representation is usually considered approximation of original signal, which can be achieved by linear superposition of relatively small number of basic elements from some predefined signal pool or a set, called dictionary, where term small usually denotes that the number of nonzero components in such approximation is significantly smaller than the dimension of original signal space. If dictionary is not known in advance, it also needs to be designed in some suitable way that will ease the main task of finding sparse signal approximation. Therefore, dictionary is also often made to be not a generative set of independent elements, but an overcomplete or redundant set, \cite{elad2010sparse}, of signals belonging to the same signal class of interest. Such approach is also considered to be an example of representation design that is based on union of subspaces, \cite{elad2010role}, as opposed to classical subspace approximation approaches. %
\par{}Optimization problems that lead to sparse representation are usually formulated as a compromise between two opposing requirements, which can be described (following corresponding formulation from \cite{olshausen1996emergence}) in the form of an energy minimization problem with criterion defined as:
\begin{equation}\label{A1}
\begin{aligned}
E = &\left\{\,\lnot{\textit{ preserved information content}}\,\right\} + \\
&\; +\; \mu \, \left\{\,\lnot \; {\textit{simplicity of representation}} \,\right\}\,,
\end{aligned}
\end{equation}
where symbol $\lnot $ denotes negation of the corresponding positive statements, while $ \mu >0$ determines the influence of the second requirement. %
\par{}Since  minimization of energy functional \eqref{A1} is not always possible in closed form, (sub)optimal solutions for dictionary learning and corresponding sparse representation of original signals are usually constructed based on an iterative procedure,  which consists of two alternating steps in each iteration that perform: a)~adaptation of the elements of the generative set with the aim of improving current sparse approximations that are utilizing  sparse codes which were already determined as a part of sparse representation of each signal in the previous iteration; b) sparse coding or representation of original signals in the generative set, i.e. dictionary, determined in the previous step of the same iteration. %
According to \cite{olshausen1997sparse},
 described strategy during the coding step removes influence of components with small contribution to preservation of original information content in constructed approximation, while the dictionary synthesis step,  through optimization of basis functions that are comprising dictionary, reduces the approximation error that results from the imposed sparsity inducing constrain. %
 \par{}Therefore, an essential part of any  sparse representation procedure is ability to properly measure corresponding approximation error, which in the case of structured signals is more complex and should be based on representation invariant quantities, i.e. preserved information content. %
 \par{}In the following lines we will show how the information distance based on Bregman matrix divergence, which was presented in Section \ref{Section3}, enables formulation of two different sparse representation problems that are used in sparse approximation of: a)~covariance descriptors; or b)~precision matrices in GMMs. In addition, it will be discussed how  the latter formulation, which is concerned with sparse representation of precision matrices, benefits from the asymptotic equivalence between ML estimation and MDI principle, which was presented in Section~\ref{Section2}, and how it can be interpreted in the same framework of information distance based on Bregman matrix divergence as the formulation used for sparse coding of covariance descriptors. Although these two sparse representation formulations have totally different applications, and as a result also motivations, it will be also pointed out how asymmetric nature of Burg divergence makes their formulations qualitatively different.
\subsection{Sparse coding of covariance descriptors}
\label{subSection4.1}
Covariance descriptors were for the first time proposed in \cite{tuzel2006region}, as fast image descriptors for texture classification and object detection in images. Motivation for their application and methods for their construction were also discussed in \cite{porikli2010method}, while the first study utilizing sparse coding of such matrix descriptors was presented in  \cite{sivalingam2010tensor}. In the context of classification, sparse coding or representation of covariance descriptors is usually used such that in the training phase each category in the recognition system is first associated with one specific dictionary, consisting of randomly chosen signal instances or learned prototypes, while in the deployment phase all class dependent dictionaries are merged together and used for sparse approximation of some new observation that comes in the form of covariance descriptor. Recognition can be based on the analysis of computed sparse code, which consists of weights used in constructed sparse approximation, counting the number of contributing components from each category dependent dictionary (followed by the subsequent category voting); or obtained codes (representation) could be just viewed as a nonlinear transformation of original features, which were initially represented by a collection of second order statistics of vector features aggregated by the corresponding covariance matrix.
\par{}The most common formulation of sparse representation of covariance descriptors can be expressed as, \cite{sivalingam2010tensor}:
\begin{equation}\label{A2}
\begin{aligned}
& \underset{{ {\rm{x}}\geq 0}    }{\mathop{ \min }}  & & D_{\mathrm {B} }\left({\rm{x}} \otimes \mathcal{A} \,,\, {{\rm{S}} }\right) \;+ \; \mu \|  {\rm{x}} \|_1  \,  \\
&\text{ s.t.}& &\; 0 \preceq {\rm{x}} \otimes \mathcal{A} \preceq  {\rm{S}}
\,,
\end{aligned}
\end{equation}
where ${\rm{A}}\succeq 0$ denotes that ${\rm{A}} \in Sym_d^{+}$,  set of positive semi-definite matrices, while ${\rm{A}}\succ 0$ corresponds to $Sym_d^{++}$, set of positive definite matrices. Similarly, ${\rm{A}} \succ {\rm{B}}$ corresponds to $({\rm{A}}-{\rm{B}}) \succ 0$, and ${\rm{A}} \succeq {\rm{B}}$ to $({\rm{A}}-{\rm{B}}) \succeq 0$. Notation $\otimes$ is introduced to concisely denote a linear combination of dictionary elements which results in sparse approximation $\hat{{\rm{S}}} = {\rm{x}} \otimes \mathcal{A} $, where vector ${\rm{x}} \in \mathbb{R}^K$ and represents the corresponding sparse code or sparse representation of original covariance  ${{\rm{S}}}$, i.e. weight coefficients of the corresponding linear combination of dictionary elements.
\par{}Symbol $\mathcal{A}$ in \eqref{A2} denotes a dictionary which consists of $K$ randomly chosen covariances from the training set, $\mathcal{A} = [ {\rm{A}}_1, \ldots {\rm{A}}_K ]$, while $\mathcal{S} = \{{\rm{S}}_j\}_{j=1}^N$ denotes a set of covariance descriptors that are approximated through sparse coding (representation) by sparse linear combination of elements from $\mathcal{A}$, which actually means that only a relatively small number of elements from $\mathcal{A}$ contributes to resulting approximation. How many is actually determined by the value of $\ell_1$ norm of vector of weight coefficients, $\|  {\rm{x}} \|_1 = \sum\nolimits_{i=1}^K x_i $, which is a convex surrogate of sparsity inducing $\ell_0$ norm.
\par{}It can be immediately noticed that \eqref{A2} resembles the energy functional in \eqref{A1}, where $\mu \geq0$ is a penalty factor that determines the influence of sparse regularizer in the given objective function, i.e. it describes the compromise between desired sparseness of the solution, on one side, and an approximation error that is captured by the Burg or LogDet matrix divergence \eqref{B_8}, on the other.%
\par{}However, since covariance descriptors can be interpreted as structured signals, such approximation also has to preserve additional signal properties, i.e.  positive definiteness of original matrices ${\rm{S}}_j$. More exactly, since  $\forall j$, ${\rm{S}}_j \in Sym_d^{++}$, each approximation $\hat{{\rm{S}}}_j $ also has to be $Sym_d^{++}$.%
\par{} In addition, presented formulation in \eqref{A2} is also specific in the sense that it assumes that $\forall i$, dictionary elements ${\rm{A}}_i \in Sym_d^{++}$, which means that sparse approximation in the form of linear combination: $\hat{{\rm{S}}}_j = \sum \nolimits_{i=1}^K{x_{ij} \, {\rm{A}}_i }$, is required to be such that unknown weight coefficients $x_{ij}$ of each approximation $\hat{{\rm{S}}}_j$ need to be always positive, which can be concisely written as ${\rm{x}}_j \geq 0$, where ${\rm{x}}_j $ is the corresponding sparse code or representation of the covariance descriptor ${\rm{S}}_j$. This is reflected in additional constraints introduced in \eqref{A2}, besides standard $0 \preceq {\rm{x}} \otimes \mathcal{A}$.
\par{}More details regarding \eqref{A2} and proposed quadratic programming solution can be found in \cite{sivalingam2014tensor} or \cite{zhang2013discriminative}, however it is important to notice that the constructed sparse approximation in \eqref{A2}  appears as the first argument of the directed Bregman matrix divergence defined in \eqref{B_8}.
\subsection{Sparse representation of precision matrices}
\label{subSection4.2}
\par{}Motivation for sparse  representation of precision matrices (inverse covariances) that appear as parameters of Gaussian mixture models (GMMs), \cite{brkljavc2014sparse}, is quite different from the one which was behind the presented sparse coding formulation in Section~\ref{subSection4.1}, and comes from the possibility to achieve more efficient use of resources in GMM based statistical classifier under some favourable conditions, like the large number of states in the corresponding system. It exploits the properties of sparse representation, which results in a vector of weight coefficients with relatively large number of zeros, by utilizing sparse approximations of originally estimated parameters of GMMs, i.e. sparse approximations of corresponding precision matrices in the system. It can be considered as an extension of the previous approaches with similar motivation, like the ones described in \cite{wang2009modeling} and \cite{huo2010precision}, which could be characterized as subspace based, and which came after \cite{gopinath1998, olsen2004modeling, axelrod2002modeling, vanhoucke2003mixtures}. On the other hand, sparse representation in  \cite{brkljavc2014sparse} can be regarded as approximation based on union of subspaces, i.e. approximation based on linear combination of overcomplete dictionary elements. In addition, the idea about exploiting sparse representation in approximation of precision matrices in GMMs came earlier in the work of \cite{jakovljevic2013}, however this approximation method was placed in a vector setting, where approximations were based on sparse representation of eigenvectors of precision matrices.
\par{}In contrast to \eqref{A2}, sparse representation of precision matrices in  \cite{brkljavc2014sparse}  has different formulation, as will be presented below. Denote original covariance matrix $j$ with ${\bar \Sigma}_j$, and its inverse (precision matrix) as ${\bar \Sigma}_j^{-1} = {\text{P}}_j $, then sparse approximation of each precision is constructed as $\hat{\text{P}}_j  = \sum_{k=1}^D \lambda_k^j \text{S}_k$, where  $\lambda_k^j$ are corresponding sparse representation coefficients, i.e. weights associated with each element of dictionary $\Psi = \left\{{\text{S}_i} \right\}_{i=1}^D$, which consists of learned symmetric matrices of size $d$. If we consider all $M$ precisions that exists in some system, set of their sparse representation coefficients, i.e. sparse codes  ${\Lambda_i} \in \mathbb{R}^D$ can be denoted as $\Lambda = \left\{ {\Lambda_i}\right\}_{i=1}^M$. Let us start from the corresponding regularized likelihood function:
\begin{equation}\label{A3}
\begin{small}
\begin{aligned}
{\cal L}\left( \Lambda,\Psi \right) ={}  \mathop \sum_{j = 1}^M  {n_j} \left\{{ {\rm tr}\bigl({\bar \Sigma}_j{\hat{\text{P}}}_j \bigr) - \ln  {\det {{\hat{\text{P}}}_j}} } \right\} {+} \mu\,  {\lVert \Lambda_j \rVert}_{1}\,,
\end{aligned}
\end{small}
\end{equation}
which is minimized under positive definiteness constraints:
\begin{equation}\label{A4}
\left( {{\Lambda}^*,{\Psi }^*} \right) = {\underset{(\Lambda , \Psi)} {\arg \min}} \: {\cal L}\left(\Lambda,\Psi \right)\;\; \text{s.t.}\;\;  {\hat{\text{P}}_j}\succ 0 \,.
\end{equation}
by an alternating iterative procedure for dictionary learning and sparse coding that was proposed in  \cite{brkljavc2014sparse}.
It can be shown that the sparse coding step of \eqref{A4} reduces to $M$ independent optimization subproblems of the form:
\begin{equation}\label{A5}
{\Lambda}^*_j = {\underset{\Lambda_j} {\arg \min}} \: {\cal L}_j\left(\Lambda_j \right)  \;\;  \text{s.t.}   \;\; \hat{\text{P}}_j\succ 0\,, \,\textrm{where:}
\end{equation}
\begin{equation}\label{A6}
{\cal L}_j\left(\Lambda_j \right) = {\rm tr}\bigl({\bar \Sigma}_j \,\hat{\text{P}}_j\bigr) - \ln {\det {\hat{\text{P}}_j} }  \: {+} \: \mu \,{\lVert \Lambda_j \rVert}_{1}\,.
\end{equation}
From \eqref{B_8} it follows that \eqref{A6} without $\ell_1$ norm is up to an additive constant equal to:
\begin{equation}\label{A7}
\begin{aligned}
 D_{\mathrm {B} }\bigl({\bar \Sigma}_j \,,\, {\hat{ \Sigma}_j}\bigr) =
 {\rm tr}\bigl({\bar \Sigma}_j \,\hat{\text{P}}_j\bigr)
- \ln \det \bigl({\bar \Sigma}_j \,\hat{\text{P}}_j\bigr)   -d
\,,
\end{aligned}
\end{equation}
since subject of optimization (sparse representation) is $\hat{\text{P}}_j$.
\par{}This confirms significance of the LogDet matrix divergence in the formulation of \eqref{A3}. However, there are two important aspects in which formulation \eqref{A6} differs from \eqref{A2}. The first one is related to the properties of dictionary elements, which in \eqref{A6} need to be only symmetric, and not positive definite. As a consequence,  sparse representation coefficients  are not restricted to be only in $\mathbb{R}^+$, which required a different approach in solving \eqref{A6} than in  \eqref{A2}. The second one is related to the asymmetric nature of the LogDet divergence \eqref{B_8}, which results in different types of approximation problems in the case of  \eqref{A6} and \eqref{A2}. In the case of covariances in \eqref{A2}, optimization is performed over the first argument of Burg divergence \eqref{B_8}, while in the case of precisions in \eqref{A6} over the second argument. However formulations are also different in the sense that in \eqref{A6} approximations are targeting precisions, i.e. reconstruction of ${{\rm{B}}}^{-1}$ from $D_{\mathrm {B} }\big({{\rm{A}} },{{\rm{B}} }\big)$ in  \eqref{B_8}, while approach in \eqref{A2} targets reconstruction of ${{\rm{A}}}$. Of course, we could write $D_{\mathrm {B} }({{\rm{B}}^{-1} }, {{\rm{A}}^{-1} })$, which would correspond to approximation of precision ${{\rm{B}}}^{-1}$  using \eqref{A2}, however it is a different kind of approximation problem, since, as discussed above, there are other constraints.
\par{}In the end, there is a question how formulation \eqref{A5}, \eqref{A6} arises from the asymptotic equivalence between ML estimation and KL divergence minimization, as discussed in Section~\ref{subSection2.3}. This can be shown, \cite{brkljac2017}, starting from the corresponding sum of Gaussian log-likelihood functions, and by using the following sample covariance identity:
\begin{equation}\label{A8}
\resizebox{0.88\hsize}{!}{$
{\mathop \sum \limits_{i = 1}^{n_j} \; n_j\; {\rm tr}\left( \frac{1}{n_j} \, {(x_{ij}-{\bar \mu}_j)^T}\,(x_{ij}-{\bar\mu}_j)\;{\hat{\text{P}}_j} \right)} \, \approx \,n_j\; {\rm tr}\bigl( {{{\bar{ \Sigma }}}_j}\,  {\hat{\text{P}}_j} \bigr) \,. $}
\end{equation}
\section{Concluding remarks}
\label{Section5}
\par{}We have considered general approaches to formulation of optimization problems of finding sparse representation of covariance
and precision matrices. Through theoretical analysis of ML estimation and Bregman divergences, and their connection with KL divergence of normal distributions, we have tried to comprehend their role in designing information preserving objectives used in sparse representation of positive definite matrices. Although we have presented two sparse representation formulations coming from two different application areas, it would be interesting to perform their experimental comparison under same application scenario in some future work.

\bibliographystyle{IEEEtran}
\bibliography{RAD_TF18}

%\begin{thebibliography}{00}
%\bibitem{sivalingam2010tensor}
%R.~Sivalingam, D.~Boley, V.~Morellas, N.~Papanikolopoulos.
%\newblock Tensor sparse coding for region covariances.
%\newblock {\em Proceedings of the 8\textsuperscript{th} European Conference on
%  Computer Vision}, pp. 722--735. Springer, 2010.
%
%\bibitem{b1} G. Eason, B. Noble, and I. N. Sneddon, ``On certain integrals of Lipschitz-Hankel type involving products of Bessel functions,'' Phil. Trans. Roy. Soc. London, vol. A247, pp. 529--551, April 1955.
%\bibitem{b2} J. Clerk Maxwell, A Treatise on Electricity and Magnetism, 3rd ed., vol. 2. Oxford: Clarendon, 1892, pp.68--73.
%\bibitem{b3} I. S. Jacobs and C. P. Bean, ``Fine particles, thin films and exchange anisotropy,'' in Magnetism, vol. III, G. T. Rado and H. Suhl, Eds. New York: Academic, 1963, pp. 271--350.
%\bibitem{b4} K. Elissa, ``Title of paper if known,'' unpublished.
%\bibitem{b5} R. Nicole, ``Title of paper with only first word capitalized,'' J. Name Stand. Abbrev., in press.
%\bibitem{b6} Y. Yorozu, M. Hirano, K. Oka, and Y. Tagawa, ``Electron spectroscopy studies on magneto-optical media and plastic substrate interface,'' IEEE Transl. J. Magn. Japan, vol. 2, pp. 740--741, August 1987 [Digests 9th Annual Conf. Magnetics Japan, p. 301, 1982].
%\bibitem{b7} M. Young, The Technical Writer's Handbook. Mill Valley, CA: University Science, 1989.
%\end{thebibliography}

%\bibliographystyle{ieeetran}
%\bibliography{ieeeabrv,mybibfile}

\end{document}